\documentclass[10pt,square,comma,twocolumn,letterpaper]{article}

\usepackage{times}  
\usepackage{subfig}
\usepackage{helvet} 
\usepackage{courier}  
\usepackage{makecell}
\usepackage{bm}
\usepackage{amsmath}
\usepackage[hyphens]{url}  
\usepackage{graphicx} 
\usepackage{caption} 
\usepackage{cite}
\usepackage{xspace}
\usepackage{amsmath}
\usepackage{array}
\usepackage{helvet}  
\usepackage{courier}
 \usepackage{mathrsfs}
\usepackage{amsmath,amssymb,amsfonts}
\usepackage{makecell}
\usepackage{multicol}
\usepackage{multirow}
\usepackage{url}
\usepackage{amsthm} %
\usepackage{algorithm}
\usepackage{algorithmic}
\usepackage{graphicx}
\usepackage{textcomp}
\usepackage{xcolor}
\usepackage{booktabs}
\usepackage{amssymb}
\usepackage{booktabs}
\usepackage{algorithm}
\usepackage{algorithmic}

\usepackage[pagebackref,breaklinks,colorlinks]{hyperref}

\usepackage[capitalize]{cleveref}
\crefname{section}{Sec.}{Secs.}
\Crefname{section}{Section}{Sections}
\Crefname{table}{Table}{Tables}
\crefname{table}{Tab.}{Tabs.}

\def\cvprPaperID{*****} 
\def\confName{****}
\def\confYear{****}

\begin{document}
\title{Efficient Stein Variational Inference for Reliable Distribution-lossless Network Pruning}
\author{Yingchun WANG\\
The Hong Kong Polytechnic University\\
Hong Kong, China\\
{\tt\small 20116342r@connect.polyu.hk}
\and
Song Guo\\
The Hong Kong Polytechnic University\\
Hong Kong, China\\
{\tt\small song.guo@polyu.edu.hk}
\and
Weizhan Zhang\\
Xi'an Jiaotong University\\
Shan Xi, China\\
\and
Yida Xu\\
Hong Kong Baptist University\\
Hong Kong, China\\
\and
Jingcai Guo\\
The Hong Kong Polytechnic University\\
Hong Kong, China\\
\and
Jie Zhang\\
The Hong Kong Polytechnic University\\
Hong Kong, China\\
}
\maketitle

\begin{abstract}
Network pruning is a promising way to generate light but accurate models and enable their deployment on resource-limited edge devices. 
However, the current state-of-the-art assumes that the effective sub-network and the other superfluous parameters in the given network share the same distribution, where pruning inevitably involves a distribution truncation operation. They usually eliminate values near zero. While simple, it may not be the most appropriate method, as effective models may naturally have many small values associated with them. Removing near-zero values already embedded in model space may significantly reduce model accuracy. Another line of work has proposed to assign discrete prior over all possible sub-structures that still rely on human-crafted prior hypotheses.
Worse still, existing methods use regularized point estimates, namely \textit{Hard Pruning}, that can not provide error estimations and fail reliability justification for the pruned networks.
In this paper, we propose a novel distribution-lossless pruning method, named DLLP, to theoretically find the pruned lottery within Bayesian treatment.
Specifically, DLLP remodels the vanilla networks as discrete priors for the latent pruned model and the other redundancy. More importantly, DLLP uses Stein Variational Inference to approach the latent prior and effectively bypasses calculating KL divergence with unknown distribution.
Extensive experiments based on small Cifar-10 and large-scaled ImageNet demonstrate that our method can obtain sparser networks with great generalization performance while providing quantified reliability for the pruned model.
%
\end{abstract}

\section{Introduction}
\label{sec:intro}
Convolutional neural networks (CNNs) have achieved great success in many visual recognition tasks such as image classification \cite{he2016deep}, object detection \cite{ren2015faster}, and image segmentation \cite{dai2016instance}, etc. The success of CNNs is inseparable from an excessive number of parameters that are well organized to perform sophisticated computations, which conflicts with the increasing demand for deploying these resource-consuming applications on resource-limited devices.\par

Network pruning has been an effective method to compress very deep neural networks without significant accuracy degradation. Guided by various heuristic regularizations, existing works identify a small enough sub-network with a deterministic point estimate that mimics the original performance. 
One of the most classic pruning methods is post-training pruning following an iterative three-stage paradigm: “training, pruning, and fine-tuning”, which is expensive and cumbersome.
Recently, initialization pruning has been proposed as a promising method to identify an effective sub-network before training based on representative gradients, greatly reducing the computational overhead of iterative training \cite{tanaka2020pruning,wang,jorge}. Note that in the following we sometimes refer to lottery tickets to refer to the pruned submodel to be solved.\par

However, both post-pruning and initialization pruning are distribution-transacted, which leads to a great accuracy drop and unavoidably brings an extra fine-tuning cost. The fact lies in that existing methods model the inherently discrete lottery and model redundancies into one prior with a continuous distribution, which implicitly makes the deletion of sub-modules affect each other. Worse still, most of these works rely on heuristic parametric hypotheses to tackle the pruning optimization towards unknown prior. These subjective prior beliefs are lacking theoretical support, which may generate misleading and unreliable results, being difficult to convince the latent users. 
Last but not least, we have to say that state-of-the-art works are overconfident about their pruned results. Hard pruning with point estimate ignores the model estimation error from both aleatoric and epistemic uncertainties, leading to low estimator efficiency and resulting in poor pruning reliability.\par 

In this work, we argue that \textit{model pruning is a problem about the parameters distribution discovery with respect to the most causal sub-module for the latent variables}. 
We propose a novel distribution-lossless pruning method (DLLP) to perform more accurate model pruning. DLLP models the true low-dimensional prior as a spike-and-slab distribution and forces the learned model to approach the true prior by Stein variational inference. Specifically, 
we explicitly assign two discrete distributions to the causal sub-module and redundant remainder.
And different from previous methods using prior hyperparameters due to the unknown true distribution, DLLP bypasses directly solving the {\rm KL} distance and uses the kernelized stein discrepancy between two distributions to calculate the gradient with respect to {\rm KL} loss. So the approximate prior distribution parameterized by all learnable parameters can be optimized efficiently by stein variational gradient descent~\cite{stein} with respect to {\rm KL} distance loss together with experience risks. After 
that, the causal sub-module/pruned model would be the slap part of the optimized spike-and-slab distribution. 
Finally, as distribution-based estimation, DLLP can naturally provide the estimation error of both model's prior and posterior, as well as the estimation efficiency. We theoretically demonstrate that DLLP significantly improves the estimation efficiency of the pruned model compared to the previous SOTA, thereby increasing the confidence of the results.
\par
In summary, our contributions are:
\begin{itemize}
\item We theoretically analyze the pruning preliminaries from the most principled Bayesian perspective,  and discover that the current pruning paradigm is distribution-truncated and relies on the heuristic hyperparameter hypothesis, inherently leading to significant accuracy reduction.
\item We propose a novel distribution-lossless pruning method, named DLLP,  to efficiently find the most accurate sub-model. The key lies in discrete prior modeling and deterministic stein variational inference.
\item We subtly bypass computing the {\rm KL} distance involving unknown distribution during pruning, and instead, we utilize the stein variational gradient to optimize along the steepest direction directly.
\end{itemize}
Extensive experiments on Cifar-10 and ImageNet datasets have demonstrated that DLLP achieves state-of-the-art pruning performance.
\section{Related Work}
\subsection{Model Pruning}
Most deep neural networks are overparameterized and unavoidably lead to expensive computing and memory overhead \cite{toornot}. Specifically, most deep networks have a large number of redundant parameters which contribute little to task performance. Previous works have demonstrated that only a small part (5-10\%) of the weights are involved in the main calculation and lots of the activation values are zero. 
Thus, model pruning has been proposed to reduce the resource cost of deep models while keeping a satisfying accuracy.\par
Early works focus on pruning deep models based on the whole dataset and sharing a compact model among all different samples \cite{wen2016learning, liu2017learning, liebenwein2019provable, molchanov2019importance, tang2020scop}. To be specific, static methods select pruning results through trade-offs on different samples, which leads to final compact models having limited representation capacity and thus suffering an obvious accuracy drop with large pruning rates. \par
Recently, some works turn their attention to the pursuit of the ultimate pruning rate and focus on excavating sample-wise model redundancy, named dynamic pruning.
Dynamic pruning generates different compact models for different samples \cite{dong2017more, gao2018dynamic, hua2019channel, rao2018runtime, tang2021manifold}. 
For example, \cite{manifold} improves the dynamic pruning efficiency by embedding the manifold information of all samples into the space of pruned networks. 
Thus, dynamic methods could achieve a higher instance-wise compression rate. However, most dynamic pruning methods require deploying the full model and performing running time path-finding during inference. Thus, dynamic pruning is not resource-efficient and less practical.
\subsection{Stein Variational Inference}
Stein discrepancy is a measurement of the distance between probability distributions defined on a metric space, which is first formulated as a method to assess the quality of Markov chain Monte Carlo samplers. Based on that, \cite{liu2016kernelized} derives a statistical method to measure the difference between complex high-dimensional probability distributions, which combines the Stein feature and the reproduced kernel Hilbert space (RKHS), named Kernelized Stein Discrepancy (KSD). Unlike MCMC and VI, KSD has unbiased statistics which help to conduct hypothesis testing. The U-statistic of KSD depends only on the score function of the known distribution and does not depend on its normalization constant. \par

Taking KSD as the optimization target of the distribution, Liu Qiang et.al. \cite{stein} proposed an approximate inference algorithm, Stein variational gradient descent (SVGD). Unlike Markov Chain Monte Carlo MCMC, it is a deterministic algorithm. 
SVGD is the natural counterpart to gradient descent applied in full Bayesian inference. Specifically, it performs (functional) gradient descent on this set of particles to minimize the ${\rm KL}$ divergence and drive the particles to fit the true posterior distribution. We take the optimization of SVGD in DLLP based on the following points:
(1) The stein operator successfully makes the unknown distribution part of the distance calculation to 0, which eliminates the problem that the unknown distribution cannot be calculated;
(2) The stein variation gives the analytical solution of the gradient direction with the fastest distribution change, which effectively increases the efficiency of optimization;

\section{Method}
In this section, we will provide a detailed illustration of the proposed \textbf{D}istribution \textbf{L}oss\textbf{L}ess \textbf{P}runing (DLLP) method and the theoretical effectiveness guarantee. In Section~\ref{sec:pre}, we first theoretically analyze previous pruning SOTAs from a principled Bayesian view, inherently revealing their essential deficiencies including distribution truncation and model unreliability. Then in Section~\ref{sec:dllp}, we discuss the two key technologies of DLLP with respect to prior distribution and stein variational inference. After that, we provide a theoretical guarantee of the model estimation efficiency of DLLP to demonstrate its effectiveness in Section~\ref{sec:the}.
\subsection{Preliminaries}\label{sec:pre}
In this subsection, we analyze several existing model pruning works from a Bayesian perspective, including the classic pruning based on $\ell_1$, $\ell_2$ regularization and weight magnitude and the cutting-edge polarization pruning based on batch normalization regularization.

\subsubsection{Model Pruning From a Bayesian Perspective}\label{sec:bay}
Given a dataset with $N$ samples as $X = \left \{ x_i \right \}_{i=1}^{N}$ with the corresponding labels $Y = \left \{ y_i \right \}_{i=1}^{N} $. We assume that the dataset follows a Gaussian distribution with variance $\delta$ and expectation $y$.
For a convolution neural network (CNN) with a parameter set $\Theta$ with $M$ items, $\omega_i$ denotes the $i_{th}$ learnable items in $\Theta$.
Assuming that the learnable model parameters follow a Laplace distribution prior with the expectation 0 and the variance $b$, we have $f(\Theta|0,b)=\frac{1}{2b}\exp^{-\frac{|\Theta|}{b}}$.
%
Then, the posterior probability of the model under the current data conditions can be defined as:
\begin{equation}
 \begin{aligned}
  p(\Theta|Y,X)&\propto p(Y|X;\Theta  ) p(\Theta ) \\
    &= \prod_{i=1}^{N}\prod_{j=1}^{M}\frac{1}{\sqrt{2\pi \delta }}  \exp^{-\frac{(y_{i}-\omega_j  x_{i})^2}{2\delta }}
   \frac{1}{2b} \exp^{-\frac{|\omega_j|}{b} },\\
\end{aligned}
\label{2}
\end{equation}   
and the log function of the conditional posterior of the referenced model is:
\begin{equation}
 \begin{aligned}    
    \log p(\Theta|Y,X)\propto
   -\sum_{i=1}^{N} \sum_{j=1}^{M} \left\{\frac{1}{2\delta }  (y_{i}-\omega_j  x_{i})^2 +\frac{1}{b}||\omega_j||_1\right\}
\end{aligned}
\label{3}
\end{equation}
Then we could get the familiar formulation of the $\ell_1$-norm regularization pruning~\cite{han2015deep}:
\begin{equation}
 \begin{aligned}    
    \Theta^* = 
     \arg \min_{\Theta }\sum_{i=1}^{N} \sum_{j=1}^{M} \left\{\frac{1}{2\delta }  (y_{i}-\omega_j  x_{i})^2 +\frac{1}{b}||\omega_j||_1\right\}
\end{aligned}
\label{4}
\end{equation}
Similarly, assuming the model prior obeys Gaussian distribution with the expectation 0 and the variance $\alpha$, the equivalent Bayesian form of $\ell_2$-norm regularization pruning can be derived as:
\begin{equation}
 \begin{aligned}    
   \Theta^* = 
     \arg \min_{\Theta }\sum_{i=1}^{N} \sum_{j=1}^{M} \left \{ \frac{1}{2\delta }(y_{i}-\omega_j   x_{i})^2+\frac{1}{2\alpha }\omega_j^2 \right\}
\end{aligned}
\label{7}
\end{equation}
Another example comes from state-of-the-art neuron-level model pruning using polarization regularizer \cite{polar}, where each neuron is associated with a scaling factor. Let $\bm{\gamma}$ represent the set of the scaling factors, and in Bayesian treatment, their $\bm{\gamma}$ follows the prior distribution as follows:
\begin{equation}
 \begin{aligned}    
p(\bm{\gamma}) = \frac{1}{2}(\left\{\frac{1}{2a}e^{-\frac{|\bm{\gamma}-\bar{\gamma }|}{a} } -\frac{1}{2a}e^{-\frac{|\bm{\gamma}|^t}{a}} \right\},
\end{aligned}
\label{8}
\end{equation}
where $\bar{\gamma}$ is the mean of all elements in $\bm{\gamma}$ and $a$ denotes the variance of Laplace distributions. The model parameters are modeled as a Gaussian distribution with the hyper-prior of $\bm{\gamma}$ as: $p(\Theta|\alpha, \bm{\gamma})=\frac{1}{\sqrt{2\pi\alpha } }e^{-\frac{(\bm{\gamma}\Theta)^2}{2\alpha}} $.
Thus, the optimization of polarization neuron pruning is equal to minimizing the negative of the following  posterior probability:
\begin{equation}
        \begin{aligned}
        &\log p(\Theta,\bm{\gamma}|Y,X) \propto p(Y,X|\Theta,\bm{\gamma})P(\Theta|\bm{\gamma})p(\bm{\gamma})\\
        &\propto-\frac{1}{2\delta }\sum_{i=1}^{N} (y_i-\Theta x_i)^2 
        -\frac{(\bm{\gamma}\Theta)^2}{2\alpha } 
       + \frac{1}{a}(t||\bm{\gamma}||-||\bm{\gamma}-\bar{\gamma } ||)
        \end{aligned}
        \label{10}
 \end{equation}
As shown in Eq.~\ref{4}, Eq.~\ref{7}, and Eq.~\ref{10}, the optimization targets of existing SOTAs are equivalent to modeling the learnable parameters as Gaussian distributions with deterministic parametric hypotheses.
However, such optimization would produce biased and inconsistent estimations for the pruned model due to the following two main limitations: 
\begin{enumerate}
    \item[(1)] \textbf{Prior sharing:} Instead of assigning discrete priors, existing SOTAs generalize all possible sub-structures into the same continuous prior, where pruning inevitably involves a distribution truncation operation. 
    Liu et.al~\cite{rethink} have pointed out that
    the well-trained convolution models should approximately follow a Gaussian-like distribution. Therefore, distribution-truncated pruning could significantly reduce model accuracy.
    We will theoretically illustrate the distributional truncation from pruning in the next subsection. 
    \item[(2)] \textbf{Parametric hypothesis:} The current SOTAs heuristically set the hyperparameters in the unknown distribution as deterministic values instead of estimable random variables. While simple, it may generate misleading results and poor generalization performance. Worse still, we theoretically demonstrate that this artificial setting could significantly reduce the estimation efficiency of the pruned model. Estimators with low efficiency bring an unreliable pruned model, especially when facing uncontrollable data domains in the real world. 
\end{enumerate}

%
%
%
\subsubsection{Distribution-truncated Pruning}\label{sec:loss}
In this subsection, we will analyze the distributions of the two parts of an overall Gaussian distribution which are torn apart by hard pruning, and explain the amount of information lost in distribution-loss pruning. \par
Let $\sigma$ be the variance of model distribution in the unpruned model and $n$ be the number of samples. Generally, the model parameters based on $\ell_1$-norm (Eq.~\ref{4}) and $\ell_2$-norm (Eq.~\ref{7}) regularization follow a \textbf{Half-normal distribution} and a \textbf{Scaled Chi distribution}, respectively. Define the power of norm terms as $p$ (1 or 2), then the distribution of the model parameters obeys ${\rm Amoroso}(0,\sqrt{2}\sigma,\frac{n(p-1)}{2},2 )$ and can be written as follows:
\begin{equation}
    \begin{aligned}
    p(\Theta)=\frac{1}{\Gamma(\frac{n(p-1)}{2})} |\frac{\sqrt{2} }{\sigma } |
    (\frac{\Theta}{\sqrt{2}\sigma } )^{n(p-1)-1}
    \exp^{-(\frac{\Theta}{\sqrt{2}\sigma } )^2}.
    \end{aligned}
    \label{11}
\end{equation}

 After training, hard pruning removes parts of estimators deemed unimportant under respective criteria. We use classical magnitude-based prune to illustrate this process. Assume $\Delta$ as the magnitude threshold. After removing ``unimportant'' weights smaller than $\Delta$, the limit distribution of the reserved tail can be modeled as a generalized Pareto distribution (GPD)~\cite{pickands1975statistical}. For a single item $w_i$ in $\Theta$, it probability density can be written as follows:
\begin{equation}
    \begin{aligned}
    p(\omega_i) = \begin{cases}
    0~~~~ \text{if}~~ |\omega_i| \le \Delta\\
   \frac{k\Delta ^k}{\omega_i^{k+1}}~~\text{else}~~ |\omega_i| > \Delta
\end{cases}
\end{aligned}
\label{12}
\end{equation}
where $k=\frac{h}{\sigma}$, $h \in \mathbb{R}$ and $h>0$. Removed nodes are part of the model prior distribution and contribute to model predictions.\\

\subsection{Detailed DLLP}\label{sec:dllp}
In this section, we will explain the implementation of DLLP from two aspects: (1) \textbf{discrete prior modeling} to overcome the biased and inconsistent estimation of the pruning model caused by the hypothetical and uniform model prior;
(2) \textbf{stein variational inference} to take the estimator uncertainties of the pruned model into consideration instead of using point estimation.
The detailed training pipeline of DLLP can be found in Algorithm.~\ref{alg1:Framwork}.
\subsubsection{Prior Distribution}
Given data likelihood:
\begin{equation}
    p(Y|X, \Theta,d)=\frac{d}{\sqrt{2\pi}}e^{\frac{d^2(Y-\Theta X)^2}{2}}
\label{eq11}
\end{equation}
where $d$ denotes the inverse of the variance of the predictions. Note that $d$ is a learnable parameter in our method.

As discussed before, DLLP expects to automatically derive the most possible causal module by distribution-lossless model pruning. Thus, it assigns different priors to the pruned model and the remaining parts. Specifically, DLLP models the prior of the referenced networks as a spike-and-slab distribution which includes two discrete Gaussian distributions combined by a Bernoulli function. The details are expressed as follows: 
%
\begin{equation}
\begin{aligned}
    &p(\Theta|\lambda, \epsilon,\Phi)\\
    &\propto \prod_{i=1}^{M}(\theta_i \frac{\lambda}{\sqrt{2\pi}}\exp^{\frac{(\lambda \omega_i)^2}{2}}  + (1-\theta_i)\frac{\epsilon}{\sqrt{2\pi}}\exp^{\frac{(\epsilon \omega_i)^2}{2}} ))
\end{aligned}
\label{eq12}
\end{equation}
where $\epsilon$ is a positive infinity and $\lim_{\epsilon \to \infty} \mathcal{N}(\omega_i|0,\frac{1}{\epsilon}) = \delta_0(\omega_i)$. $\lambda \in \mathbb{R+}$ is a learnable parameter. $\theta_{i}$ is the probability for the Bernoulli distribution. For each variable $\omega_i$, it belongs to the most possible causal sub-module with probability $\theta_{i}$ and obeys the gaussian distribution $\mathcal{N}(0,\frac{1}{\lambda})$. Instead, it follows $\mathcal{N}(0,\frac{1}{\epsilon})$  with probability $1-\theta_i$. And $\Phi$ is the set of $\theta$. 

Then we could derive the weights posterior as follows:
\begin{equation}
    \begin{aligned}
&p(\Theta |Y,X) \propto p(Y|X, \Theta,d ) p(\Theta|\lambda,\epsilon, \Phi)\\
& = \prod_{i=1}^N\prod_{j=1}^M\mathcal{N}(\omega_j x_i,d) \left\{\theta_j \mathcal{N}(0,\frac{1}{\lambda})+(1-\theta_j)  \mathcal{N}(0,\frac{1}{\epsilon})\right\}
\end{aligned}
\label{18}
\end{equation}
In the following, we may abbreviate $p(\Theta |Y,X)$ as $p_{\Theta}$. Assuming a variational distribution $q(\Theta)$ ($q_{\Theta}$) to approximate the above posterior, the  optimization of the model distribution during sparsity training can be constructed as follows:
\begin{equation}
    \begin{aligned}
q^*(\Theta) = \arg~\min_{q} \max_p {\rm KL}\left\{q(\Theta)||p(\Theta|Y,X)\right\}
\end{aligned}
\label{20}
\end{equation}

\subsubsection{Optimization Objective}
DLLP is defined as a hybrid optimization with respect to the minimization of task loss and distribution distance. The overall objective is defined as:
\begin{equation}
    \begin{aligned}
    &\min \mathcal{J}(\Theta) = \min \mathbb{E}_{X} \mathbb{E}_{\Phi} L_{\Theta}(X)\\
    &= \min \mathbb{E}_X \mathbb{E}_{\Phi}
    \left[ \mathcal{L}(f_\Theta(X), Y) + \beta {\rm KL} (q(\Theta)||p(\Theta|Y,X)) \right]
    \end{aligned}
    \label{15}
\end{equation}
where $f_{\Theta}(X)$ denotes the model predictions for input $X$, and $\beta$ is a tuning hyperparameter to adjust the weights of the two optimization terms.

Note that the model prior is ambiguous with unknown normalization parameters, so it is hard to compute the accurate distance between $q(\Theta)$ and the target $p(\Theta|Y, X)$. Unlike previous work that uses heuristic assumptions to force the calculation of ${\rm KL}$~\cite{efficient, variational}, we cleverly bypass solving ${\rm KL}$ with unknown distributions and instead seek the optimal direction that makes the ${\rm KL}$ decrease the fastest. 
Specifically, we regard each model updating as a smooth transformation of $q(\Theta)$, and our goal is to find out an optimal transformation function corresponding to the maximum drop of the ${\rm KL}$ function.
Define functional that maps all possible transformation directions of $q(\Theta)$ to the declines in the ${\rm KL}(q(\Theta)||p(\Theta|Y, X))$. It has been proved that the derivative direction of the functional is the negative of the kernelized stein discrepancy between the distributions~\cite{stein}, which can be described as:
\begin{equation}
\setlength{\abovedisplayskip}{12pt}
\setlength{\belowdisplayskip}{12pt}
\begin{aligned}
       &-S(q(\Theta||p(\Theta|Y,X)))\\
    &=-\mathbb{E}_{\Theta \sim q}\left[ k(\Theta,\cdot )\nabla_{\Theta}\log p(\Theta|Y,X)+\nabla_{\Theta}k(\Theta,\cdot)     \right] 
\end{aligned}
\label{eq16}
\end{equation}
where $k$ denotes a kernel function, which is a radial basis function kernel in DLLP.

To optimized the objective in Eq.~\ref{15}, we can derive the gradients with respect to $\Theta$ as follows:
\begin{equation}
\setlength{\abovedisplayskip}{12pt}
\setlength{\belowdisplayskip}{12pt}
    \begin{aligned}
  &\nabla _{\Theta}\mathcal{J(\Theta)}=\mathbb{E}_{\rm X} \mathbb{E}_{\rm \Phi} 
\nabla _{\Theta} L(\Theta,X)\\
&=\mathbb{E}_{\rm X} \mathbb{E}_{\rm \Phi}  \nabla_{\Theta} \mathcal{L}(f_{\Theta},Y)-\beta  S(q(\Theta||p(\Theta|Y,X)))\\
    \end{aligned}
    \label{17}
\end{equation}
The first part in Eq.~\ref{17} is the cross entropy loss of the task predictions while the other two parts correspond to the KL loss between the current model distribution and the target posterior. Note that we use the continuous Gumble-Softmax trick to sample possible weights from the discrete Bernoulli binomial distribution so that Eq.~\ref{15} can be optimized end-to-end in a gradient descent manner. 
And DLLP scales down the temperature of the gumble softmax during the training process to gradually approach the real discrete distribution.

\begin{algorithm}[htb] 
\caption{Network pruning based on discrete prior modeling and efficient stein variational gradient descent.} 
\label{alg1:Framwork} 
\begin{algorithmic}[1] 
\REQUIRE ~~\\ 
A randomly initialized referenced model with learnable parameters set $M_0 = \left\{\Theta_0, d_0,\lambda_0, \left\{\theta_i\right\}^0 \right\}$;\\
Another twin model with $M_1= \left\{\Theta_1, d_1,\lambda_1,\left\{\theta_i\right\}^1 \right\}$;\\
A target distribution $p(\Theta)$ as in Eq. \ref{18}.
\ENSURE ~~\\ 
The referenced model $M_0$ that approximates $p(\Theta)$;\\
The pruned subpart $M_p$;
\REPEAT
\STATE $K(\Theta_0,\Theta_1) = \exp(-\frac{1}{\log 2}||\Theta_0-\Theta_1||_2^2) \leftarrow$ Compute the kernelized model distance;
\STATE $(x_i,y_i) \leftarrow$ Randomly select a data batch;
\FOR{$m \in [M_0, M_1]$}
\STATE $\hat{y}_i \leftarrow$ Perform a model forward propagation ;
\STATE $p(\Theta|y_i,x_i) \leftarrow$ Compute posterior probability based on Eq. \ref{eq11}, Eq. \ref{eq12} and Eq. \ref{18};
\STATE $\Delta m \leftarrow$ Compute model gradient based on Eq. \ref{eq16} and Eq. \ref{17};
\STATE $m \leftarrow$ Update model with gradient $\Delta m$
\ENDFOR
\UNTIL convergence of $M_0$
\STATE $M_p \leftarrow$ Extract the slab part from $M_0$ which follows a spike-and-slab distribution;
\RETURN $M_0,M_p$
\end{algorithmic}
\end{algorithm}

\subsection{Theoretical Guarantee}\label{sec:the}
In this subsection, we will illustrate that the prior hypothesis and point estimation in the current SOTA severely reduces the estimation efficiency of pruned estimators, resulting in untrustworthy lightweight models. We first theoretically compute the ideal Cramer-Rao lower bound (CRLB) for the pruned estimators. Based on that, we provide the estimation efficiency of the current SOTA in four cases. The results show that existing pruning paradigms have poor credibility, thus being 
impractical in the noisy real world.

Given the set of data samples $X$, assume that the outputs of multiple forward passes obey the observation equation: $Y_k=\Theta \top X+n_k, k=1, 2, ..., N$, wherein $K$ represents the number of observations, and $n_k$ is an independent and identically distributed Gaussian random noise with mean as 0 and variance as $\epsilon^2$. 
The estimator $\Theta$ is a Gaussian random variable with mean as $\mu_{\Theta}$ and variance as $\alpha^2$, also, $\Theta \top X \sim \mathcal{N}(\mu_{\Theta} \top X, \alpha^2)$. Since $X$ is a constant, in the following we would use $\Theta$ and $\Theta \top X$ interchangeably. Based on the optimization target defined in Eq. \ref{18}, Eq. \ref{20} and Eq. \ref{15}, the maximum posterior estimator of $\Theta$ can be computed as:
\begin{equation}
    \hat{\Theta}_{map} = \frac{\bar{Y} \alpha^2+\mu_{\Theta}\epsilon^2} {\epsilon^2+\alpha^2}
\end{equation}
where $\bar{Y}=\frac{1}{K} \sum_{k=1}^{K}Y_k$. As in SOTAs which use the point estimator for pruned models, the variances of model parameters $\alpha^2$ are confidently set to zero, and we have $\mathbb{E}[\hat{\Theta}_{map}]=\mathbb{E}[\Theta]=\mu_{\Theta}$. Under this condition, $\hat{\Theta}_{map}$ is an unbiased estimate of $\Theta$, and $\mathbb{E}[\hat{\Theta}_{map}\top X - \Theta \top X] = \sum_{k=1}^{\infty}  (Y_k - \Theta \top X) p(Y_k|X, \Theta) =0$.

Thus, for a pruned model, the ideal CRLB of its unbiased estimator could be calculated as:
\begin{equation}
    \mathbb{E}[\hat{\Theta}_{map}\top X - \Theta \top X] \ge 
    \frac{1}{-\mathbb{E}[\frac{\partial^2 \ln p(Y_k|X, \Theta)}{\partial \Theta^2}] }
\end{equation}
We note the lower bound of the pruned estimator as ${\rm CRLB}(\hat{\Theta})=\epsilon^2$.

Next, we discuss the estimation efficiency of the pruned model in four real-world cases, to illustrate the poor unreliability of pruning SOTAs.\\
\textbf{X, $\Theta$ without noise:} This ideal situation is the same as the setup in SOTA above. The estimator variance could be calculated as: ${\rm Var}(\hat{\Theta})={\rm Var}(Y_k)= \epsilon^2$, and the conditional estimation efficiency would be: ${\rm e}(\Theta)=\frac{{\rm CRLB}(\hat{\Theta})}{var(\hat{\Theta}_{map})}=1$. \\
In fact, the point estimator of the pruned network is fully efficient if and only if neither the model training process nor the data distribution is noisy. \\
\textbf{X without noise, $\Theta$ with noise:} Assume there is a model uncertainty and $\alpha^2 \neq 0$. According to ~\cite{what}, ${\rm Var}(\hat{\Theta}_{map})\approx \epsilon^2+\alpha^2$, and its estimation efficiency is: ${\rm e}(\Theta)=\frac{{\rm CRLB}(\hat{\Theta})}{{\rm Var} (\hat{\Theta}_{map})}=\frac{\epsilon^2}{\epsilon^2+\alpha^2}<1$.\\
\textbf{X with noise, $\Theta$ without noise:} Similarly, assume $p(Y|X) \sim \mathcal{N}(0, \beta^2)$, the estimation efficiency would be: ${\rm e}(\Theta)=\frac{{\rm CRLB}(\hat{\Theta})}{{\rm Var}(\hat{\Theta}_{map})}=\frac{\epsilon^2}{\epsilon^2+\beta^2}<1$.\\
\textbf{X with noise, $\Theta$ with noise:} With the same distribution assumptions, the estimation efficiency of the pruned model is: ${\rm e}(\Theta)=\frac{{\rm CRLB}(\hat{\Theta})}{{\rm Var} (\hat{\Theta}_{map})}=\frac{\epsilon^2}{\epsilon^2+\alpha^2+\beta^2}<1$.

It can be explicitly seen that as noise (data or model) increases, the validity of the pruned estimators could significantly decrease under the previous frame. The proposed DLLP can naturally improve the estimation efficiencies for both the pruned model and task prediction. At the same time, it also provides estimation errors, significantly increasing the reliability of the pruned networks. The relevant proofs of the formulas will be shown in the supplementary material.

\section{Experiment}
In this section, we perform extensive experiments to analyze and validate the effectiveness of the proposed DLLP. We first describe the experimental setup of DLLP in Section.~\ref{sec:im}. Then, to study the effectiveness of the indirect KL distance optimization to both model distribution and task accuracy, we conduct ablation studies of the stein kernelized discrepancy based on the spike-and-slab distribution in Section~\ref{sec:abl}. Finally, in Section~\ref{sec:com}, we compare the proposed DLLP with state-of-the-art model pruning methods with respect to both task performance and model reliability. More details and experiment results will be shown in Supplemental Materials.

\subsection{Implement Details}\label{sec:im}
\subsubsection{Baseline}
\textbf{Model:} We explore the effectiveness of our distribution-lossless pruning method on various popular network architectures. For VGG with plain architecture, we choose the widely-used VGG-16 BN with interspersed batch normalization layers. For ResNet with residual structure, we comprehensively select Bottleneck-style ResNet-50 and Basic-style ResNet-56. 
Despite the challenges, such extensive comparison more effectively confirms the broad applicability of the proposed DLLP.

Note that these networks are improved accordingly to DLLP. For example, we add a global variable for each network as a learnable parameter to capture the noise from the uncontrollable observation. Also, we add a learnable local variable for each convolution module to model the insufficient model learning. Finally, each convolution parameter is associated with a Bernoulli variable to make a discrete choice about which of all possible sub-modules the parameter belongs to. To make the distribution sampling process differentiable, we use the continuous Gumble Softmax trick in practice. 
\\

\textbf{Dataset:} Some works have pointed out that the same pruning method may perform differently on datasets with different sample numbers. Thus, we use various popular image datasets in model pruning, including the small CIFAR-10 dataset and large-scaled ImageNet. Specifically, Cifar-10 contains only 60,000 nature color images uniformly distributed in 10 classes. While ImageNet, currently the largest database for image recognition, is more difficult to be predicted. It contains more than 1.2 million natural images and corresponds to 1000 categories.

\subsubsection{DLLP Training}
As discussed earlier, DLLP naturally induces the most causally pruned sub-modules in a one-shot training process by assigning different prior distributions to all possible sub-modules. The training process is conducted on 2 Nvidia GTX 3090TI GPUs. Each architecture is trained for 60 epochs to observe long-term effects. The batch size is set as 512 for all training. We start from the learning rate of 0.1 and gradually reduce 
it to 0.001. The only hyperparameter in our optimization objective is $\beta$ which is empirically set as 0.1. \par
After optimization, we visualized the model distribution results under all architectures. Specifically, the pruned model derived by DLLP is a discrete sub-distribution from the original model distribution, while the pruned model obtained by all other methods is the sparse original model itself.

\subsection{Ablation Study}\label{sec:abl}
\subsubsection{Pruned Model Distribution}
An important property of DLLP is that after pruning, DLLP should observe a smooth Gaussian distribution rather than a destroyed incomplete distribution. This phenomenon stems from the fact that the pruned model obtained by DLLP is an independent and most causal sub-module in the original model, and there is no correlation between all possible sub-modules of the model. 
The pruning model obtained by other SOTA methods is the residual result after heuristically savagely deleting a part of the original model. All possible sub-modules caused by model pruning are highly correlated because of the same origin.

\begin{figure}
\centering
\subfloat{
    \includegraphics[width=1.5in]{ 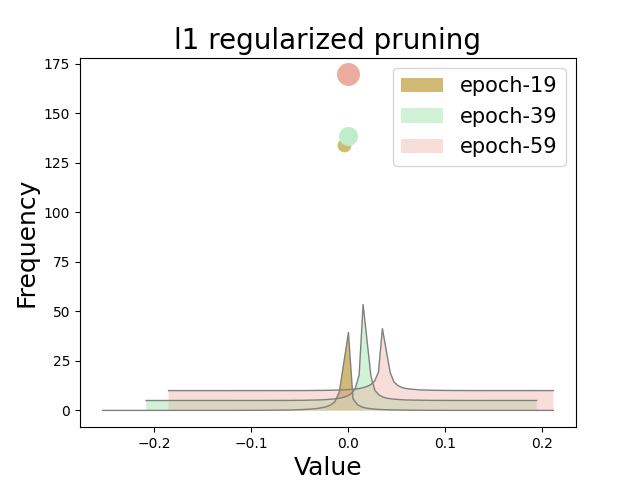}}
\subfloat{
		\includegraphics[width =1.5in]{ 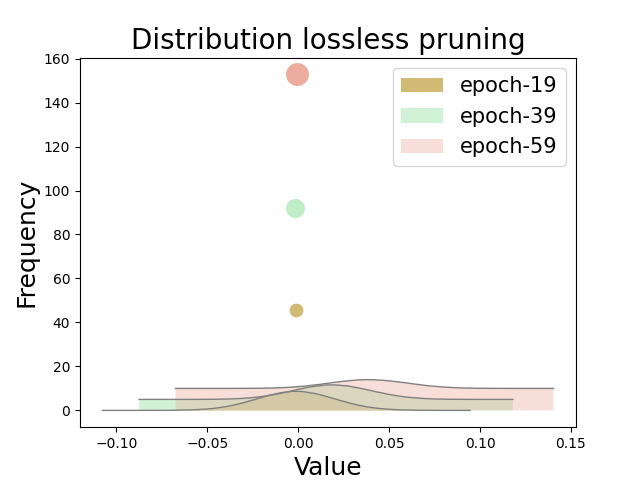}}
	\caption{The comparison of layer-wise model distributions between l1-regularized pruning and DLLP. The models are sparsely trained before actual pruning. The vertical axis is the result of the bin's raw count divided by the total number of counts and the bin width.  To magnify the distribution difference, the bin nearest to zero is extracted and reflected in the colored circle. The larger the circle, the larger the count. Note that since DLLP has a very high spike distribution at zero, so the value of the corresponding circle has been subtracted by 225 for better presentation.}
	\label{fig5}
 \vspace{-0.2cm}
\end{figure}

To demonstrate the above analysis, we use DLLP and the most typical magnitude-based pruning methods to slim VGG-11 and count their parameter distributions. 
The figures which statistically visualize the parameter distributions of the original referenced model, the pruned model of DLLP, and the pruned models of magnitude-based SOTA including typical $l_1$ and $l_2$ regularization can be found in Appendix.
Specifically, the distribution of the reference model has been proved to roughly obey the Gaussian distribution~\cite{rethink}, which is consistent with our experiments. The results also show that typical magnitude-based pruning explicitly destroys the trained model distribution, even with sparsity regularization. Their pruned distributions are bimodal distributions and have to rely on the process of fine-tuning to return to the original distribution family. Finally, for the pruned model from DLLP, as expected, the pruned distribution still obeys a smooth Gaussian distribution, which not only saves the extra fine-tuning effort but also minimizes the distribution loss.
Another strong piece of evidence comes from Figure~\ref{fig5}. The l1-regular sparsely trained model roughly follows the Laplace distribution, which is consistent with its prior. And as expected, the model after DLLP sparse training obeys the spike-and-slab distribution, which helps achieve the proposed distribution-lossless pruning.

\subsubsection{Pruned Model Reliability}
Model pruning reduces the fit of the model parameters to the data and thus may lead to higher model uncertainty. Therefore, it is critical to analyze the task uncertainty of pruned models that will be deployed in edge environments. To the best of our knowledge, we are the first work that can simultaneously account for model uncertainty while model pruning.

In this subsection we examine two types of uncertainty that DLLP can cope with: Aleatoric uncertainty refers to the noise inherent in the observation sample/training dataset, and epistemic uncertainty captures the model errors from the training \cite{what}. 

\textbf{Aleatoric Uncertainty.}
Define a function that maps from data quality to model aleatoric uncertainty. 
More specifically, the prediction results could be unreliable for samples with large noise, and the function value should be large.
Conversely, for data with better quality (such as clear or prominent foreground samples), the model is more confident in its prediction results, and the value of prediction uncertainty should be smaller.

\begin{figure}
\centering
\subfloat{
    \includegraphics[width=1.5in]{ 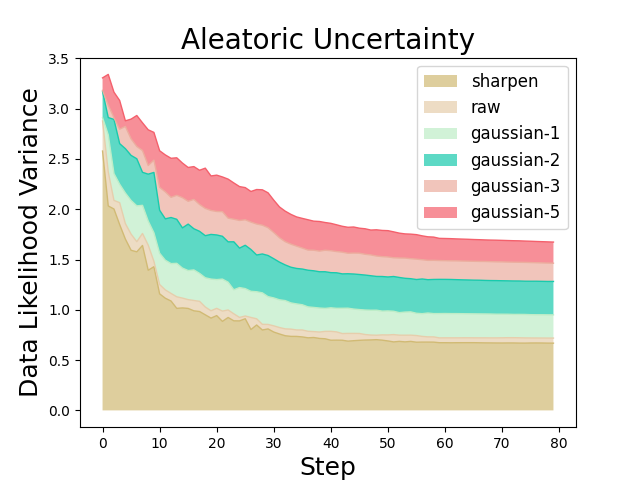}
    	\label{fig3a}}
\subfloat{
        \includegraphics[width=1.5in]{ 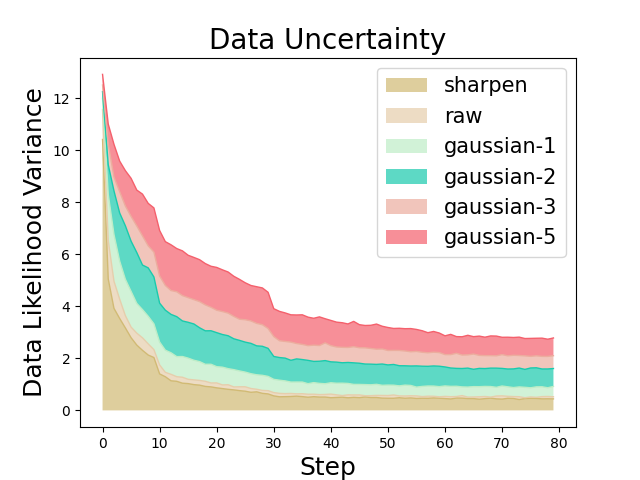}
        \label{fig3b}}
	\caption{Model uncertainty reported by DLLP under five different data situations.}
	\label{fig3}
\end{figure}

In order to demonstrate the above analysis, we conduct experiments with respect to the aleatoric uncertainty of VGG-11 predicting on processed Cifar-10 datasets.
We apply Gaussian blur matrices with the radius of 1, 2, 3, and 5 pixels to the Cifar-10 dataset to reduce the high-frequency components of the image and add data noise with different levels. 
As analyzed in Section~\ref{sec:bay}, we model the aleatoric uncertainty as the prediction variance in the likelihood function of the pruned models.
The results are shown in Figure~\ref{fig3a}.

Among them, sharpen is the data processed by Laplace sharpening, which is used to represent high-quality samples; gau-1, gau-2, gau-3, and gau-5 are Gaussian blurred data with radii of 1, 2, 3, and 5, respectively, used to represent different degrees of low-quality samples.
As expected, as the quality of the data decreases, the arbitrary uncertainty of the predictions of the pruned model rises.

\textbf{Epistemic Uncertainty}
Epistemic Uncertainty gives a measure of the extent to which the pruned sub-module is the causal model for the seen data domain. It comes from insufficient model training or insufficient observation samples. Define a function that maps from data quality to model epistemic uncertainty. Since the insufficient model training could be easily solved, the problem of insufficient observation would be our focus.
To be specific, model predictions should be less reliable if there is not sufficient full data in the training domain, and the function value should be large. Conversely, if there are sufficient observations in the training domain during pruning, the retained sub-model will be more confident in its predictive ability, and the uncertainty function will be lower.

To demonstrate the above analysis, we test the epistemic uncertainty of pruned models on VGG-11 and Cifar-10 dataset. We randomly sample 20\%, 40\%, 60\%, and 80\% of the training set and check the prediction variance reported by the likelihood function. The results are shown in Figure~\ref{fig3b}. As expected, the model epistemic uncertainty given by DLLP increases with decreasing observational data.

In summary, DLLP builds the model posterior distribution from a Bayesian perspective and sets the agnostic normalization parameters as learnable parameters. Therefore, DLLP can provide uncertainty measures for the current pruned sub-structures, and make this lightweight model more reliable. 

\subsection{Comparison with SOTA}\label{sec:com}
In this section, we compare the proposed DLLP with various SOTA methods including $l_1$ regularized pruning \cite{ns,l1}, mixed $l_1$ and $l_2$ regularized pruning \cite{deephoyer}, polarization regularized pruning \cite{c1}, reinforcement learning based pruning \cite{amc}, etc. 
The performance comparison between different pruning methods is in terms of the model accuracy and FLOPs reduction. 
\begin{table}[!tp]
\centering
\caption{Performance comparison of pruning performance on ResNet-56 and Cifar-10.}
\begin{tabular}{lccc}
\toprule
\multirow{2}{*}{Method} & Original & \multicolumn{2}{c}{Pruned} \\ 
\cmidrule{2-2}\cmidrule{3-4}
                        & Acc. (\%) & $\downarrow$FLOPs & $\downarrow$Acc. (\%) \\ 
\midrule \midrule
L1-norm \cite{l1} & 93.42         & 34\%   & 0.5  \\  
AMC \cite{amc}     & 92.80            & 50\%   & 0.9 \\  
DCP\cite{dcp}         & 93.80          & 50\%   & 0.31 \\   
NS \cite{ns}         & 93.80         & 48\%   & 0.53 \\   
SFP \cite{sfp}         & 93.59         & 51\%   & 0.24 \\   
CCP \cite{CCP}         & 93.50          & 47\%   & 0.04 \\   
FPGM \cite{fpgm}        & 93.59       & 53\%   & 0.1 \\   
Deephoyer \cite{deephoyer}        & 93.80   & 48\% & 0.26 \\  
\midrule
\textbf{DLLP} & \textbf{92.80}  & \textbf{55\%}  & \textbf{0.04}       \\  \bottomrule                                                
\end{tabular}
\label{table1}
\end{table}

\begin{table}[!tp]
\centering
\caption{Performance comparison of pruning performance on VGG-16 and Cifar-10.}
\begin{tabular}{lccc}
\bottomrule
\multirow{2}{*}{Method} &Original & \multicolumn{2}{c}{Pruned} \\ 
\cmidrule{2-2}\cmidrule{3-4}
                        & Acc. (\%) & $\downarrow$FLOPs & $\downarrow$Acc. (\%) \\ 
\midrule \midrule
LIWS \cite{liws}     &  93.45          & 70.6\%   & 0.8 \\  
ThiNet \cite{thinet}         & 93.36         & 50\%   & 0.14 \\   
CP \cite{cp}        & 93.18       & 50\%   & 0.32 \\   
RNP \cite{rnp}        & 92.15   & 50\% & 0.85 \\ 
FBS \cite{fbs}        & 93.03   & 50\% & 0.47 \\  
\midrule
\textbf{DLLP} & \textbf{92.80}   & \textbf{74.65\%}  & \textbf{1.24}        \\  
\bottomrule  
\end{tabular}
\label{table2}
\end{table}

\begin{table}[!tp]
\centering
\caption{Performance comparison of pruning performance on ResNet-50 and ImageNet.}
\begin{tabular}{lccc}
\bottomrule
\multirow{2}{*}{Method} &Original & \multicolumn{2}{c}{Pruned} \\ 
\cmidrule{2-2}\cmidrule{3-4}
                        & Acc. (\%) & $\downarrow$FLOPs & $\downarrow$Acc. (\%) \\ 
\midrule \midrule
NS \cite{ns} & 76.15         & 53\%   & 1.27 \\  
SSS \cite{sss}     & 76.12           & 43\%   & 4.30 \\  
DCP \cite{dcp}         & 76.01          & 56\%   & 1.06 \\  
CCP \cite{CCP}        & 76.15           & 54\%   & 0.94 \\   
SFP \cite{sfp}        & 76.15          & 62.14\% & 14.01 \\  
\midrule
\textbf{DLLP} & \textbf{76.15}  & \textbf{65.95\%}  & \textbf{6.13}        \\  \bottomrule                                                 
\end{tabular}
\label{table3}
\end{table}

The experiment results on Cifar-10 dataset with ResNet-56 and VGG-16 have been shown in Table~\ref{table1} and Table~\ref{table2}, respectively. It can be seen that DLLP achieves the best performance on all structures including residual style and plain style. Specifically, as shown in Table. \ref{table1}, DLLP achieves the smallest accuracy drop (0.04\%) and the highest FLOPs reduction (55\%) on ResNet-56 network, and the corresponding values are 1.24\%, 74.65\% for VGG-16 BN, which far exceeds the model compression ratio of existing SOTA.

The experiment results on ImageNet with ResNet-50 have been shown in Table~\ref{table3}. It can be seen that on a larger and more complicated dataset, DLLP still achieves the most significant FLOPs reduction (65.95\%) with an accuracy drop close to SOTA's. In summary, extensive experiments on Cifar-10 and ImageNet datasets have demonstrated that DLLP achieves state-of-the-art pruning performance.

\section{Conclusion}
In this paper, we analyze model pruning in principled Bayesian treatment and propose a novel distribution-lossless pruning method (DLLP). DLLP models the reference model as a spike-and-slab distribution to achieve distribution-lossless structure removal. What's more, DLLP uses Stein Variational Inference to force the pruned distribution to gradually approach the true prior in an efficient gradient descent manner. Even better, DLLP can quantitatively provide a measure of the prediction uncertainty for the current pruning model.
Extensive experiments have shown that compared to pruning SOTA, DLLP achieves a higher pruning rate along with satisfying accuracy.

{\small
\bibliographystyle{ieee_fullname}
\bibliography{cvpr2023-author_kit-v1_1-1/latex/PaperForReview}
}

\end{document}


\maketitle

\section{A. Proofs For Section.~3}

\begin{proof}[Proof of \textbf{Equation.~16}]
Given the set of data samples $X$, assume that the outputs of multiple forward passes obey the observation equation: $Y_k=\Theta \top X+n_k, k=1, 2, ..., N$, wherein $K$ represents the number of observations, and $n_k$ is an independent and identically distributed Gaussian random noise with mean as 0 and variance as $\epsilon^2$. We use $Y=\left\{Y_k\right\}_{k=1}^K$ to represent the set of observations. 
The estimator $\Theta$ is a Gaussian random variable with mean as $\mu_{\Theta}$ and variance as $\alpha^2$, also, $\Theta \top X \sim \mathcal{N}(\mu_{\Theta} \top X, \alpha^2)$. 
Since $X$ is a constant, in the following we would use $\Theta$ and $\Theta \top X$ interchangeably. Based on the optimization of posterior-based stein variational inference, the maximum posterior estimator of $\Theta$ can be computed as:
\begin{equation}
\begin{aligned}
    &p(\hat{\Theta}|Y,X) \propto p(Y|X, \Theta) p(\Theta)= 
    \prod_{k=1}^{K} \frac{1}{\sqrt{2 \pi \epsilon^2}} \exp^{-\frac{Y_k-\mu_{\Theta} X}{2 \epsilon^2}}
    \frac{1}{\sqrt{2 \pi \alpha^2}} \exp^{-\frac{\Theta-\mu_{\Theta}}{2 \alpha^2}} \\   
    & \Rightarrow ~~~~~ \log p(\hat{\Theta}|Y,X) \propto \frac{1}{K}\sum_{k=1}^{K} -\frac{(Y_k-\Theta \top X)^2}{2 \epsilon^2} -\frac{(\Theta - \mu_{\Theta})^2}{2 \alpha^2}\\    
\end{aligned}
\end{equation}
Let $\frac{\partial \log p(\hat{\Theta}|Y,X)}{\partial \Theta} = 0$, and we have:
\begin{equation}
\begin{aligned}
    \frac{1}{K}\sum_{k=1}^{K} \frac{(Y_k-\Theta \top X)^2}{\epsilon^2} - \frac{\Theta - \mu_{\Theta}}{\alpha^2} = 0   
\end{aligned}
\end{equation}
, and then, we can derive:
\begin{equation}
\begin{aligned}
    \hat{\Theta}_{map} = \frac{\bar{Y} \alpha^2+\mu_{\Theta}\epsilon^2}{\epsilon^2+\alpha^2}  
\end{aligned}
\end{equation}
where $\bar{Y}=\frac{1}{K}\sum_{k=1}^{K}$. 
\end{proof}
\begin{proof}[Proof of \textbf{Equation.~17}]
In state-of-the-art paradigms which use point estimator for the pruned model, the variance of model parameters $\alpha^2$ is confidently set to zero, where $\mathbb{E}[\hat{\Theta}_{map}]=\mathbb{E}[\Theta]=\mu_{\Theta}$. Under this condition, $\hat{\Theta}_{map}$ is an unbiased estimate of $\Theta$. Thus, It can be derived that: 
\begin{equation}
    \begin{aligned}
&\mathbb{E}[\hat{\Theta}_{map}\top X - \Theta \top X]  = 
 \frac{1}{K}\sum_{k=1}^{K} (Y_k- \Theta \top X)p(Y_k|X, \Theta) =0  
\end{aligned}
\end{equation}
\begin{equation}
    \begin{aligned}
& \frac{1}{K}\sum_{k=1}^{K} \frac{\partial (Y_k-\Theta \top X)p(Y_k|X, \Theta) }{\partial \Theta} =  \frac{1}{K}\sum_{k=1}^{K} p(Y_k|X, \Theta) + \frac{1}{K}\sum_{k=1}^{K} (Y_k-\Theta^T X)\frac{\partial p(Y_k|X, \Theta)}{\partial \Theta} =0\\
\end{aligned}
\end{equation}
Since $ \frac{1}{K}\sum_{k=1}^{K}p(Y_k|X, \Theta)=1$,
\begin{equation}
    \begin{aligned}
\left\{ \frac{1}{K}\sum_{k=1}^{K}(Y_k-\Theta^T X) \frac{\partial \ln p(Y_k|X, \Theta)}{\partial \Theta}  p(Y_k|X, \Theta) \right\}^2=1
\end{aligned}
\end{equation}
Based on Cauchy-Schwartz inequality, we have:
\begin{equation}
    \begin{aligned}
    &\left\{ \frac{1}{K}\sum_{k=1}^{K} (Y_k-\Theta^T X) \frac{\partial \ln p(Y_k|X, \Theta)}{\partial \Theta}  p(Y_k|X, \Theta)  \right\}^2\\
    &\le\frac{1}{K}\sum_{k=1}^{K} (Y_k-\Theta^T X)^2 p(Y_k|\Theta,x ) \frac{1}{K}\sum_{k=1}^{K}\left \{\frac{\partial \ln p(Y_k|X, \Theta)}{\partial \Theta } \right\}^2 p(Y_k|X, \Theta) \\
    \end{aligned}
\end{equation}
In the form of expectations, the above equation is equal to:
\begin{equation}
    \begin{aligned}
    &\mathbb{E}[( \hat{\Theta}_{map}  \top X-\Theta^T X)^2] \mathbb{E}\left\{\left ( \frac{\partial \ln p(\bar{Y}|X, \Theta)}{\partial \Theta } \right )  ^2\right \}\ge 1\\
    \Rightarrow &\mathbb{E}\left[( \hat{\Theta}_{map}  \top X-\Theta^T X)^2\right ] 
    \ge \frac{1}{\mathbb{E}\left\{\left ( \frac{\partial \ln p(\bar{Y}|X, \Theta)}{\partial \Theta} \right )  ^2\right \}}\\
    \end{aligned}
\end{equation}
Since the following formula holds,
\begin{equation}
    \begin{aligned}
        \mathbb{E}\left\{\left ( \frac{\partial \ln p(\bar{Y}|X, \Theta)}{\partial \Theta} \right )  ^2\right \} = 
        -\mathbb{E}\left\{\left ( \frac{\partial^2 \ln p(\bar{Y}|X, \Theta)}{\partial^2 \Theta } \right )  \right \}
    \end{aligned}
\end{equation}
Thus, for a pruned model, the ideal CRLB of its unbiased estimator could be calculated as:
\begin{equation}
    \mathbb{E}[\hat{\Theta}_{map}\top X - \Theta \top X] \ge 
    \frac{1}{-\mathbb{E}[\frac{\partial^2 \ln p(\bar{Y}|X, \Theta)}{\partial \Theta^2}] }=\epsilon^2
\end{equation}
We note the lower bound of the pruned estimator as ${\rm CRLB}(\hat{\Theta})=\epsilon^2$.
\end{proof}

\section{B.Visualization}{\label{sec:b}}
\subsection{B.1 Model Distributions}\label{B.1}
After pruning, DLLP should observe a smooth Gaussian distribution rather than a destroyed incomplete distribution. While for pruned models, due to the uniform prior, the distribution of the pruned model and the distribution of residual parameters will probably not be complete before fine-tuning.

\begin{figure*}[htp]
\centering
\subfloat[Referenced distribution]{
    \includegraphics[width=1.6in]{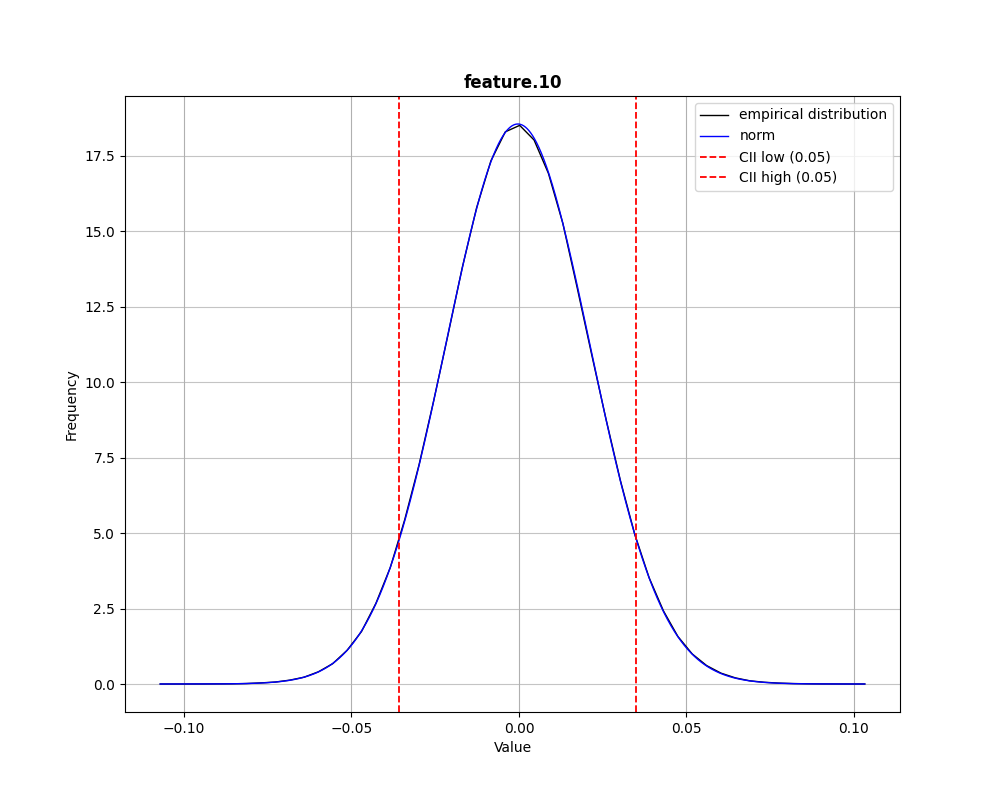}\label{a}}
\subfloat[$l_1$ pruned distribution]{
        \includegraphics[width=1.6in]{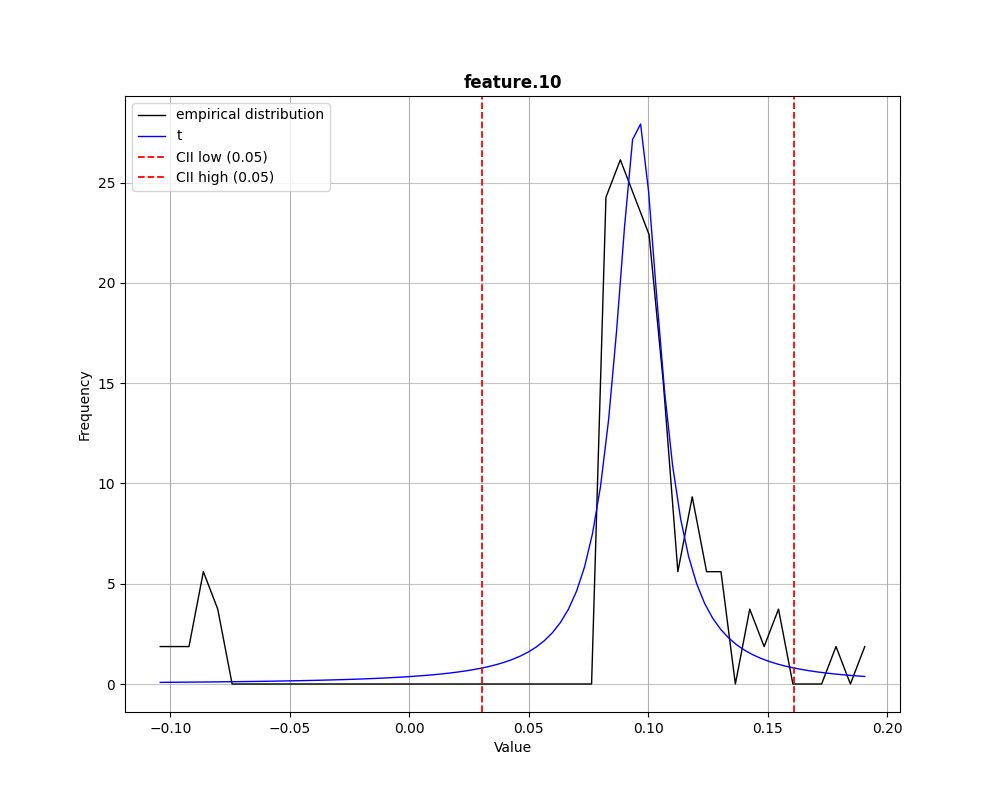}\label{b}}
\subfloat[$l_2$ pruned distribution]{
		\includegraphics[width =1.6in]{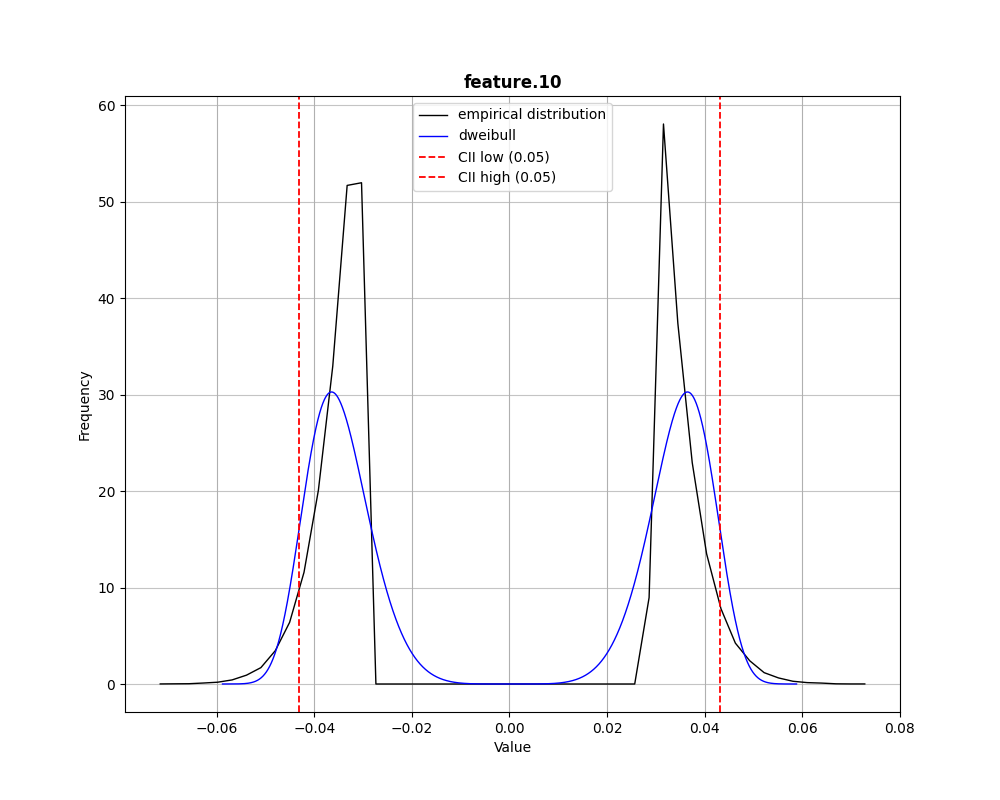}\label{c}}
\subfloat[DLLP pruned distribution]{
		\includegraphics[width =1.6in]{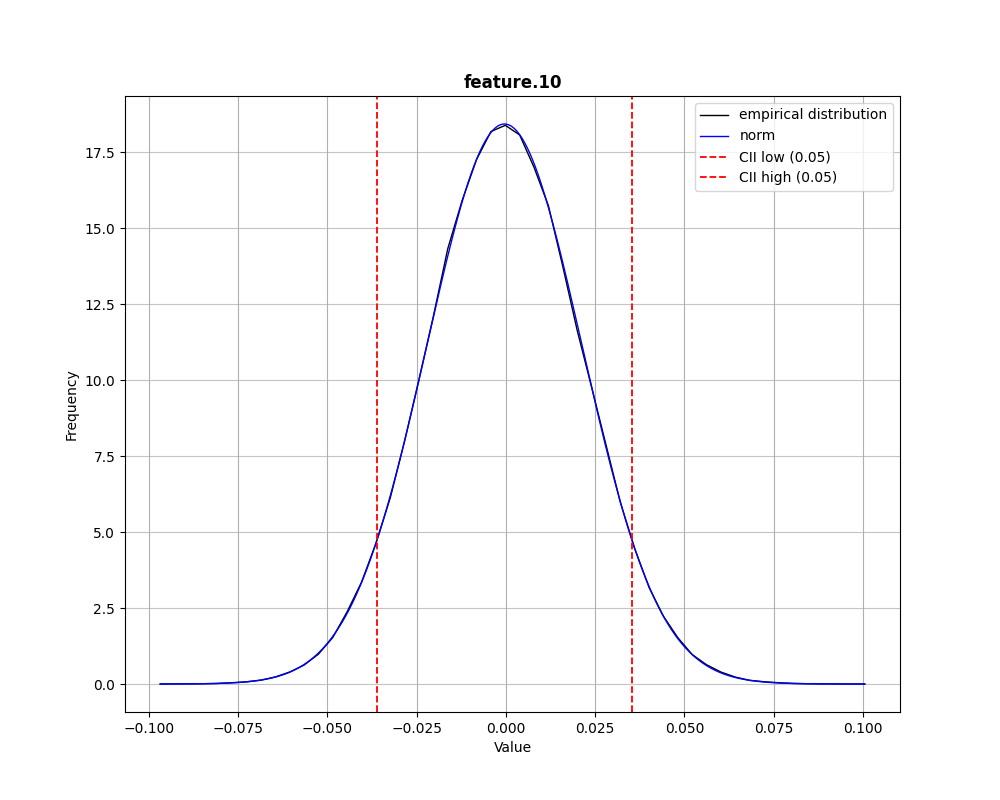}\label{d}}
	\caption{The distribution analysis of the original model, the pruned model from magnitude-based method and the pruned model from DLLP.}
	\label{fig2}
\end{figure*}

To intuitively demonstrate the above analysis, we use DLLP and the most typical magnitude-based pruning methods to slim VGG-11 and count their parameter distributions. 
As shown in Figure. \ref{fig2}, we statistically visualize the parameters distributions of the original referenced model, the pruned model before fine-tuning of DLLP, and the pruned models before fine-tuning of magnitude-based SOTA including typical $l_1$ and $l_2$ regularization. 

The pruned distribution results visually verify our above analysis.

\begin{figure}
\centering
\subfloat[Model likelihood function]{
    \includegraphics[width=0.25\columnwidth]{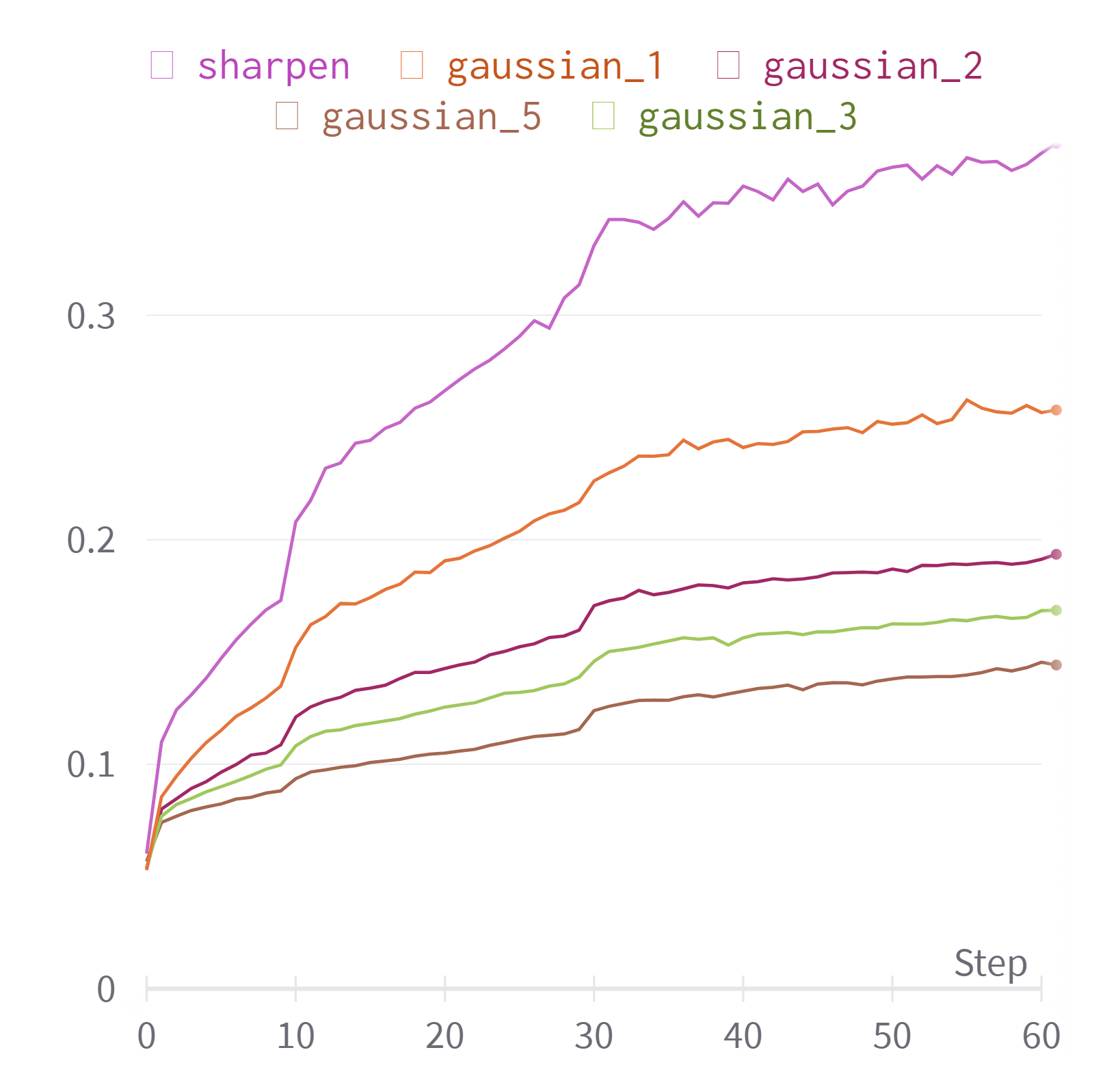}
    \label{fig4a}}
\subfloat[Model posterior probability]{
        \includegraphics[width=0.25\columnwidth]{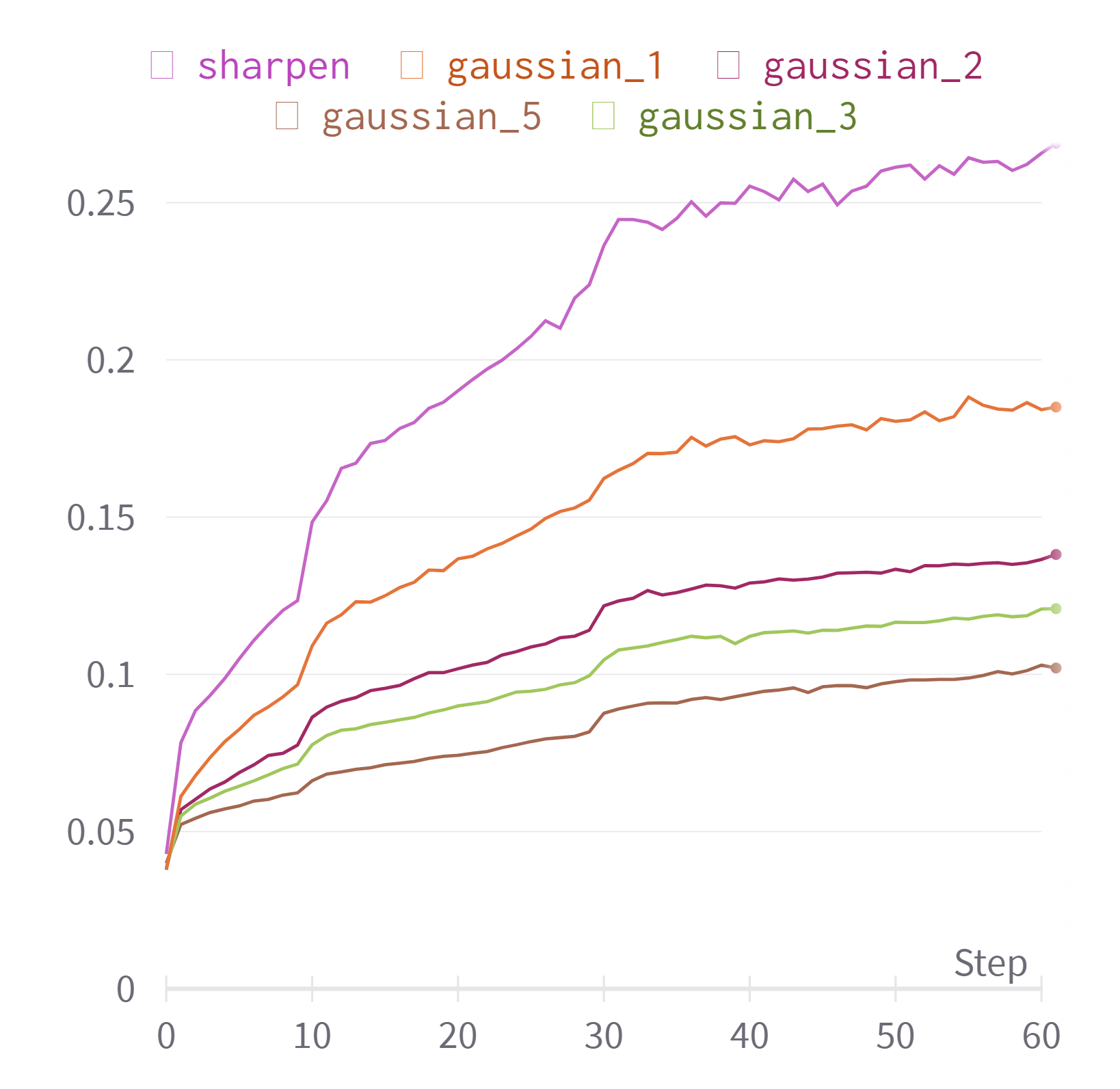}
    \label{fig4b}}
\subfloat[Model predicted variance in likelihood]{
    \includegraphics[width=0.25\columnwidth]{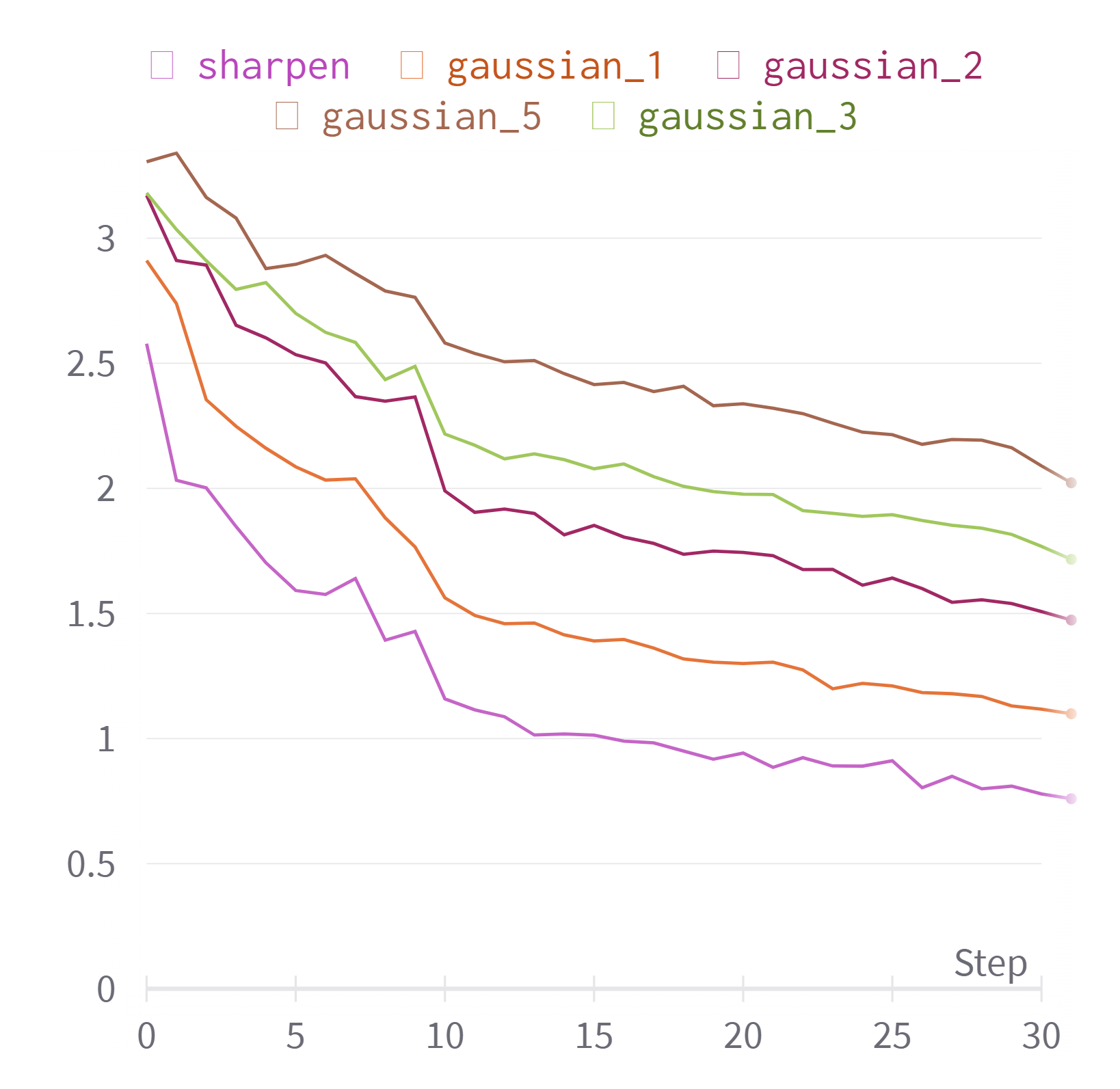}
    	\label{fig3a}}
\subfloat[Model predicted variance in posterior]{
        \includegraphics[width=0.25\columnwidth]{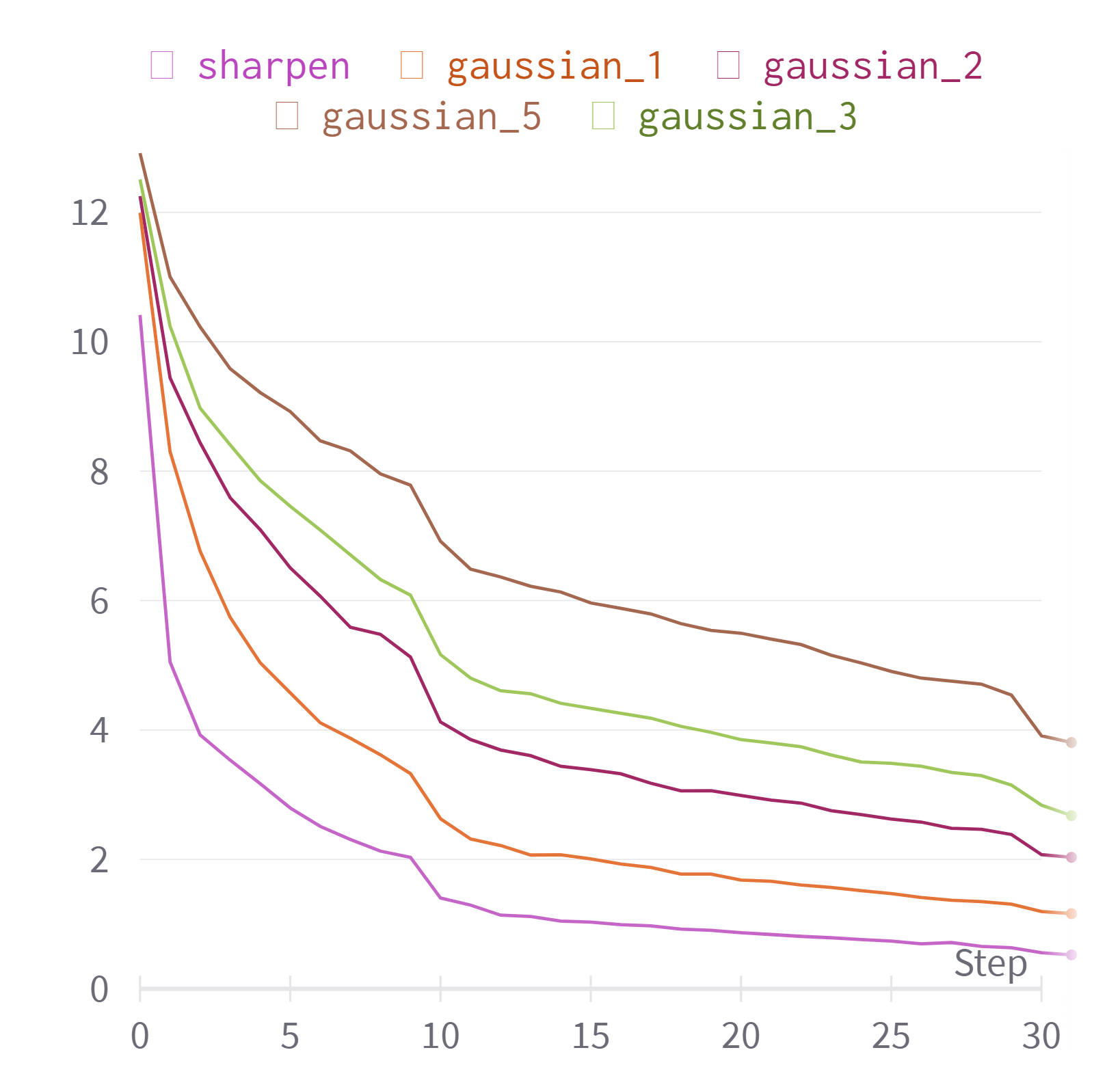}
        \label{fig3b}}
	\caption{Model prediction likelihood and posterior probability reported by DLLP under five different data qualities.}
	\label{fig4}
\end{figure}

To demonstrate the effectiveness of DLLP from the most principled Bayesian perspective, we also visualize the likelihood function which capture the aleatoric uncertainty inherently in the observations, and also the posterior probability of the model which reflects the uncertainty of the currently learned parameters. The results are shown in Figure. \ref{fig4}. The experiments are conducted on Cifar-10 dataset and VGG-11 architecture.

In our experiment setting, we use the biased data samples, which has a static one-to-one digit-color relationship, as the ID data samples. On the contrary, the OOD data samples have randomly assigned colored backgrounds. To further demonstrate that our method can successfully prune spurious features regardless of the mapping relations, we evaluate the test accuracy in 5 different mapping relation settings and draw the mean accuracy in Figure~\ref{shuffle}. As the experiment result shows, SFP's mean accuracy is relatively higher, and the variance of the accuracy is relatively lower, which shows the superiority of the proposed SFP is stable and robust. This suggests that SFP successfully prunes the sub-network associated with any spurious feature.